# Iterative Optimization and Simplification of Hierarchical Clusterings

**Doug Fisher**                                                   DFISHER@VUSE.VANDERBILT.EDU
*Department of Computer Science, Box 1679, Station B*
*Vanderbilt University, Nashville, TN 37235 USA*

## Abstract

Clustering is often used for discovering structure in data. Clustering systems differ in the *objective function* used to evaluate clustering quality and the *control strategy* used to search the space of clusterings. Ideally, the search strategy should consistently construct clusterings of high quality, but be computationally inexpensive as well. In general, we cannot have it both ways, but we can partition the search so that a system inexpensively constructs a 'tentative' clustering for initial examination, followed by iterative optimization, which continues to search in background for improved clusterings. Given this motivation, we evaluate an inexpensive strategy for creating initial clusterings, coupled with several control strategies for *iterative optimization*, each of which repeatedly modifies an initial clustering in search of a better one. One of these methods appears novel as an iterative optimization strategy in clustering contexts. Once a clustering has been constructed it is judged by analysts – often according to task-specific criteria. Several authors have abstracted these criteria and posited a generic performance task akin to pattern completion, where the error rate over completed patterns is used to 'externally' judge clustering utility. Given this performance task, we adapt resampling-based pruning strategies used by supervised learning systems to the task of simplifying hierarchical clusterings, thus promising to ease post-clustering analysis. Finally, we propose a number of objective functions, based on attribute-selection measures for decision-tree induction, that might perform well on the error rate and simplicity dimensions.

## 1. Introduction

Clustering is often used for discovering structure in data. Clustering systems differ in the *objective function* used to evaluate clustering quality and the *control strategy* used to search the space of clusterings. Ideally, the search strategy should consistently construct clusterings of high quality, but be computationally inexpensive as well. Given the combinatorial complexity of the general clustering problem, a search strategy cannot be both computationally inexpensive and give any guarantee about the quality of discovered clusterings across a diverse set of domains and objective functions. However, we can partition the search so that an initial clustering is inexpensively constructed, followed by iterative optimization procedures that continue to search in background for improved clusterings. This allows an analyst to get an early indication of the possible presence and form of structure in data, but search can continue as long as it seems worthwhile. This seems to be a primary motivation behind the design of systems such as AUTOCLASS (Cheeseman, Kelly, Self, Stutz, Taylor, & Freeman, 1988) and SNOB (Wallace & Dowe, 1994).

This paper describes and evaluates three strategies for iterative optimization, one inspired by the iterative 'seed' selection strategy of CLUSTER/2 (Michalski & Stepp, 1983a, 1983b), one is a common form of optimization that iteratively reclassifies single observations, and a third method appears novel in the clustering literature. This latter strategy was inspired, in part, by macro-learning strategies (Iba, 1989) – collections of observations are reclassified *en masse*, which appears to mitigate problems associated with local maxima as measured by the objective function. For evaluation purposes, we couple these strategies with a simple, inexpensive procedure used by COBWEB (Fisher, 1987a, 1987b) and a system by Anderson and Matessa (1991), which constructs an initial hierarchical clustering. These iterative optimization strategies, however, can be paired with other methods for constructing initial clusterings.

Once a clustering has been constructed it is judged by analysts – often according to task-specific criteria. Several authors (Fisher, 1987a, 1987b; Cheeseman et al., 1988; Anderson & Matessa, 1991) have abstracted these criteria into a generic performance task akin to pattern completion, where the error rate over completed patterns can be used to 'externally' judge the utility of a clustering. In each of these systems, the objective function has been selected with this performance task in mind. Given this performance task we adapt resampling-based pruning strategies used by supervised learning systems to the task of simplifying hierarchical clusterings, thus easing post-clustering analysis. Experimental evidence suggests that hierarchical clusterings can be greatly simplified with no increase in pattern-completion error rate.

Our experiments with clustering simplification suggest 'external' criteria of simplicity and classification cost, in addition to pattern-completion error rate, for judging the relative merits of differing objective functions in clustering. We suggest several objective functions that are adaptations of selection measures used in supervised, decision-tree induction, which may do well on the dimensions of simplicity and error rate.

## 2. Generating Hierarchical Clusterings

Clustering is a form of unsupervised learning that partitions observations into classes or clusters (collectively, called a clustering). An objective function or quality measure guides this search, ideally for a clustering that is optimal as measured by the objective function. A hierarchical-clustering system creates a tree-structured clustering, where sibling clusters partition the observations covered by their common parent. This section briefly summarizes a simple strategy, called *hierarchical sorting*, for creating hierarchical clusterings.

### 2.1 An Objective Function

We assume that an observation is a vector of nominal values, $V_{ij}$ along distinct variables, $A_i$. A measure of *category utility* (Gluck & Corter, 1985; Corter & Gluck, 1992),

$$CU(C_k) = P(C_k) \sum_i \sum_j [P(A_i = V_{ij}|C_k)^2 - P(A_i = V_{ij})^2],$$

and/or variants have been used extensively by a system known as COBWEB (Fisher, 1987a) and many related systems (Gennari, Langley, & Fisher, 1989; McKusick & Thompson, 1990; Iba & Gennari, 1991; McKusick & Langley, 1991; Reich & Fenves, 1991; Biswas, Weinberg, & Li, 1994; De Alte Da Veiga, 1994; Kilander, 1994; Ketterlin, Gançarski, & Korczak, 1995). This measure rewards clusters, $C_k$, that increase the *predictability* of variable values within $C_k$ (i.e., $P(A_i = V_{ij}|C_k)$) relative to their predictability in the population as a whole (i.e., $P(A_i = V_{ij})$). By favoring clusters that increase predictability (i.e., $P(A_i = V_{ij}|C_k) > P(A_i = V_{ij})$), we also necessarily favor clusters that increase variable value *predictiveness* (i.e., $P(C_k|A_i = V_{ij}) > P(C_k)$).

Clusters for which many variable values are predictable are *cohesive*. Increases in predictability stem from the shared variable values of observations within a cluster. A cluster is well-separated or *decoupled* from other clusters if many variable values are predictive of the cluster. High predictiveness stems from the differences in the variable values shared by members of one cluster from those shared by observations of another cluster. A general principle of clustering is to increase the similarity of observations within clusters (i.e., cohesion) and to decrease the similarity of observations across clusters (i.e., coupling).

Category utility is similar in form to the *Gini Index*, which has been used in supervised systems that construct decision trees (Mingers, 1989b; Weiss & Kulikowski, 1991). The Gini Index is typically intended to address the issue of how well the values of a variable, $A_i$, predict *a priori* known class labels in a supervised context. The summation over Gini Indices reflected in $CU$ addresses the extent that a cluster predicts the values of all the variables. $CU$ rewards clusters, $C_k$, that most reduce a collective impurity over all variables.

In Fisher's (1987a) COBWEB system, $CU$ is used to measure the quality of a partition of data, $PU(\{C_1, C_2, \ldots C_N\}) = \sum_k CU(C_k)/N$ or the average category utility of clusters in the partition. Sections 3.5 and 5.2 note some nonoptimalities with this measure of partition quality, and suggest some alternatives. Nonetheless, this measure is commonly used, we will take this opportunity to note its problems, and none of the techniques that we describe is tied to this measure.

## 2.2 The Structure of Clusters

As in COBWEB, AUTOCLASS (Cheeseman et al., 1988), and other systems (Anderson & Matessa, 1991), we will assume that clusters, $C_k$, are described probabilistically: each variable value has an associated conditional probability, $P(A_i = V_{ij}|C_k)$, which reflects the proportion of observations in $C_k$ that exhibit the value, $V_{ij}$, along variable $A_i$. In fact, each variable value is actually associated with the number of observations in the cluster having that value; probabilities are computed 'on demand' for purposes of evaluation.

Probabilistically-described clusters arranged in a tree form a hierarchical clustering known as a probabilistic categorization tree. Each set of sibling clusters partitions the observations covered by the common parent. There is a single *root* cluster, identical in structure to other clusters, but covering all observations and containing frequency information necessary to compute $P(A_i = V_{ij})$'s as required by category utility. Figure 1 gives an example of a probablistic categorization tree (i.e., a hierarchical clustering) in which each node is a cluster of observations summarized probabilistically. Observations are at leaves and are described by three variables: `Size`, `Color`, and `Shape`.

## 2.3 Hierarchical Sorting

Our strategy for initial clustering is *sorting*, which is a term adapted from a psychological task that requires subjects to perform roughly the same procedure that we describe here (Ahn & Medin, 1989). Given an observation and a current partition, sorting evaluates the quality of new clusterings that result from placing the observation in each of the existing clusters, and the quality of the clustering that results from creating a new cluster that only covers the new observation; the option that yields the highest quality score (e.g., using $PU$) is selected. The clustering grows incrementally as new observations are added.

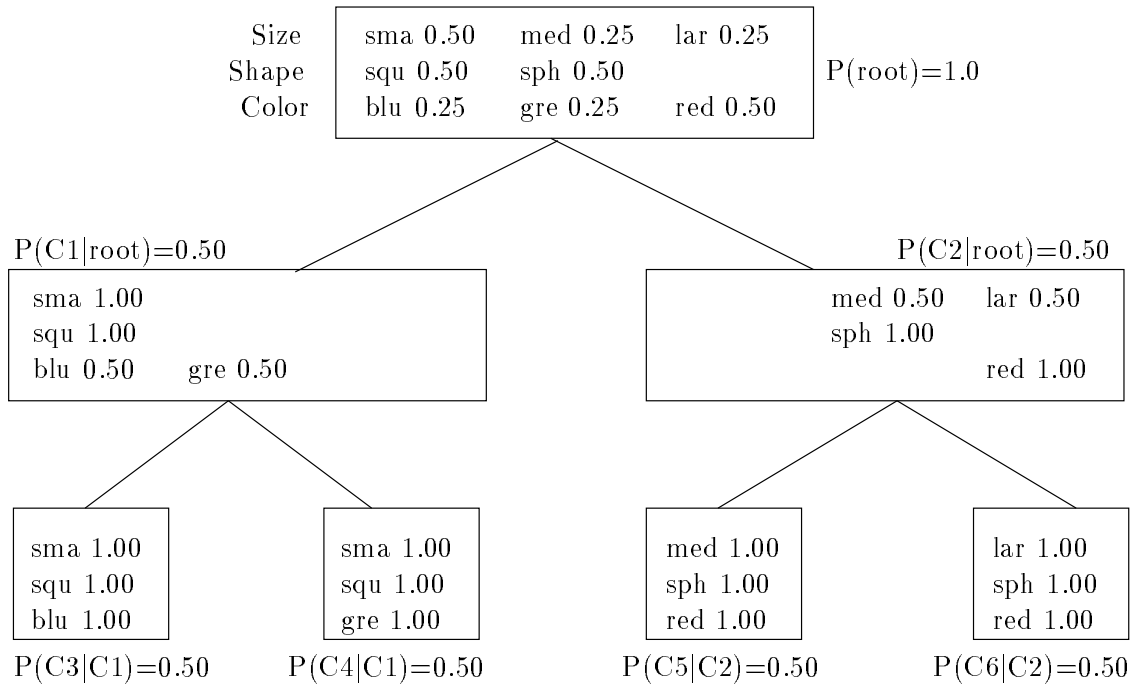

Figure 1: A probabilistic categorization tree.

This procedure is easily incorporated into a recursive loop that builds tree-structured clusterings: given an existing hierarchical clustering, an observation is sorted relative to the top-level partition (i.e., children of the root); if an existing child of the root is chosen to include the observation, then the observation is sorted relative to the children of this node, which now serves as the root in this recursive call. If a leaf is reached, the tree is extended downward. The maximum height of the tree can be bounded, thus limiting downward growth to fixed depth. Figure 2 shows the tree of Figure 1 after two new observations have been added to it: one observation extends the left subtree downward, while the second is made a new leaf at the deepest, existing level of the right subtree.

This sorting strategy is identical to that used by Anderson and Matessa (1991). The children of each cluster partition the observations that are covered by their parent, though the measure, $PU$, used to guide sorting differs from that of Anderson and Matessa. The observations themselves are stored as singleton clusters at leaves of the tree. Other hierarchical-sort based strategies augment this basic procedure in a manner described in Section 3.3 (Fisher, 1987a; Hadzikadic & Yun, 1989; Decaestecker, 1991).

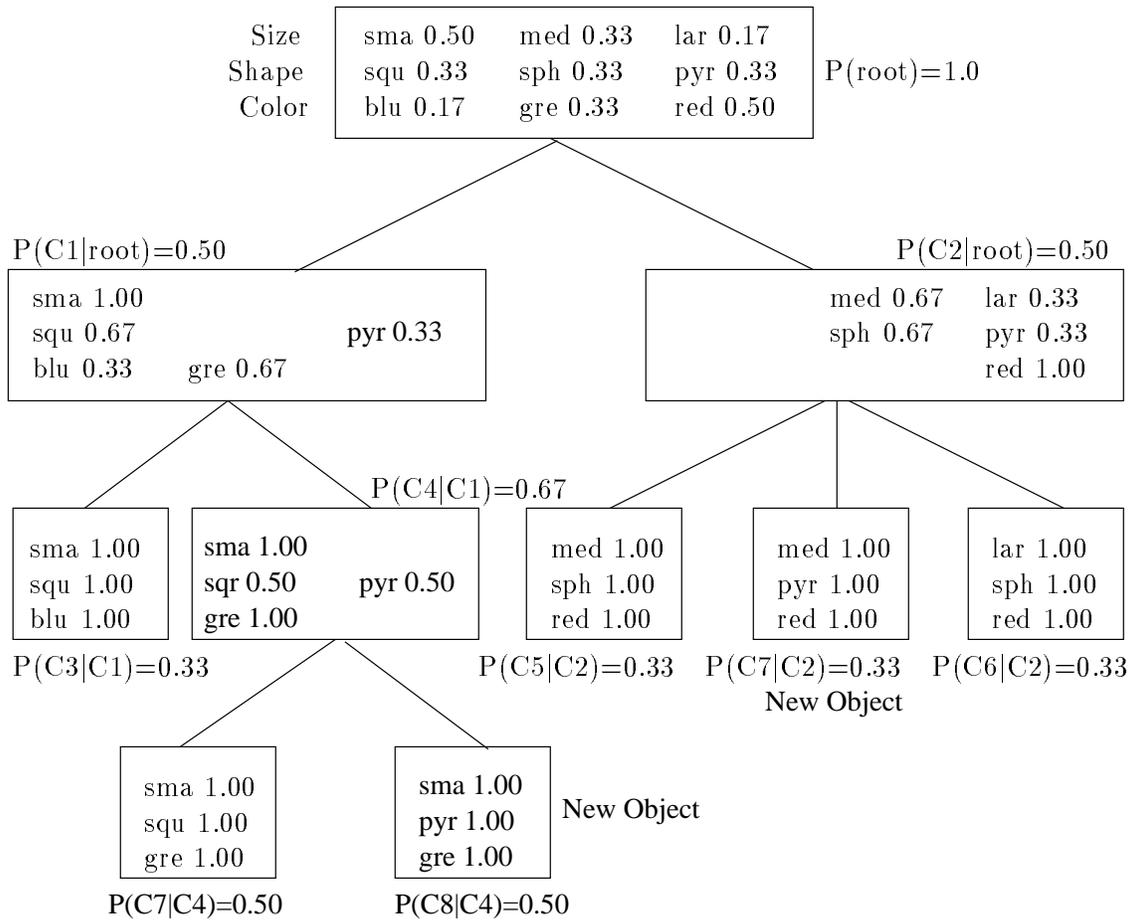

Figure 2: An updated probabilistic categorization tree.

## 3. Iterative Optimization

Hierarchical sorting quickly constructs a tree-structured clustering, but one which is typically nonoptimal. In particular, this control strategy suffers from *ordering* effects: different orderings of the observations may yield different clusterings (Fisher, Xu, & Zard, 1992). Thus, after an initial clustering phase, a (possibly offline) process of iterative optimization seeks to uncover better clusterings.

### 3.1 Seed Selection, Reordering, and Reclustering

Michalski and Stepp's (1983a) CLUSTER/2 seeks the optimal K-partitioning of data. The first step selects K random 'seed' observations from the data. These seeds are 'attractors' around which the K clusters are grown from the remaining data. Since seed selection can greatly impact clustering quality, CLUSTER/2 selects K new seeds that are 'centroids' of the K initial clusters. Clustering is repeated with these new seeds. This process iterates until there is no further improvement in the quality of generated clusterings.

```
Function ORDER(Root)
   If Root is a leaf Then Return(observations covered by Root)
      Else Order children of Root from those covering the most
              observations to those covering the least.
         For each child, $C_k$, of Root (in order) Do $L_k \leftarrow$ ORDER($C_k$)
         $L \leftarrow$ MERGE($\{L_k|$list of objects constructed by ORDER($C_k$)$\}$)
         Return($L$)
```

Table 1: A procedure for creating a 'dissimilarity' ordering of data.

Ordering effects in sorting are related to effects that arise due to differing fixed-K seed selections: the initial observations in an ordering establish initial clusters that 'attract' the remaining observations. In general, sorting performs better if the initial observations are from diverse areas of the observation-description space, since this facilitates the establishment of initial clusters that reflect these different areas. Fisher, Xu, and Zard (1992) showed that ordering data so that consecutive observations were dissimilar based on Euclidean distance led to good clusterings. Biswas *et al.* (1994) adapted this technique in their ITERATE system with similar results. In both cases, sorting used the $PU$ score described previously.

This procedure presumes that observations that appear dissimilar by Euclidean distance tend to be placed in different clusters using the objective function. Taking the lead from CLUSTER/2, a measure-independent idea first sorts using a random data ordering, then extracts a biased 'dissimilarity' ordering from the hierarchical clustering, and sorts again. The function of Table 1 outlines the reordering procedure. It recursively extracts a list of observations from the most probable (i.e., largest) cluster to the least probable, and then merges (i.e., interleaves) these lists, before exiting each recursive call – at each step, an element from the most probable cluster is placed first, followed by an element of the second most probable, and so forth. Whatever measure guides clustering, observations in differing clusters have been judged dissimilar by the measure. Thus, this measure-independent procedure returns a measure-dependent dissimilarity ordering by placing observations from different clusters back-to-back.

Following initial sorting, we extract a dissimilarity ordering, recluster, and iterate, until there is no further improvement in clustering quality.

### 3.2 Iterative Redistribution of Single Observations

A common and long-known form of iterative optimization moves single observations from cluster to cluster in search of a better clustering (Duda & Hart, 1973). The basic strategy has been used in one form or another by numerous sort-based algorithms as well (Fisher et al., 1992). The idea behind iterative *redistribution* (Biswas, Weinberg, Yang, & Koller, 1991) is simple: observations in a single-level clustering are 'removed' from their original cluster and resorted relative to the clustering. If a cluster contains only one observation, then the cluster is 'removed' and its single observation is resorted. This process continues until two consecutive iterations yield the same clustering.

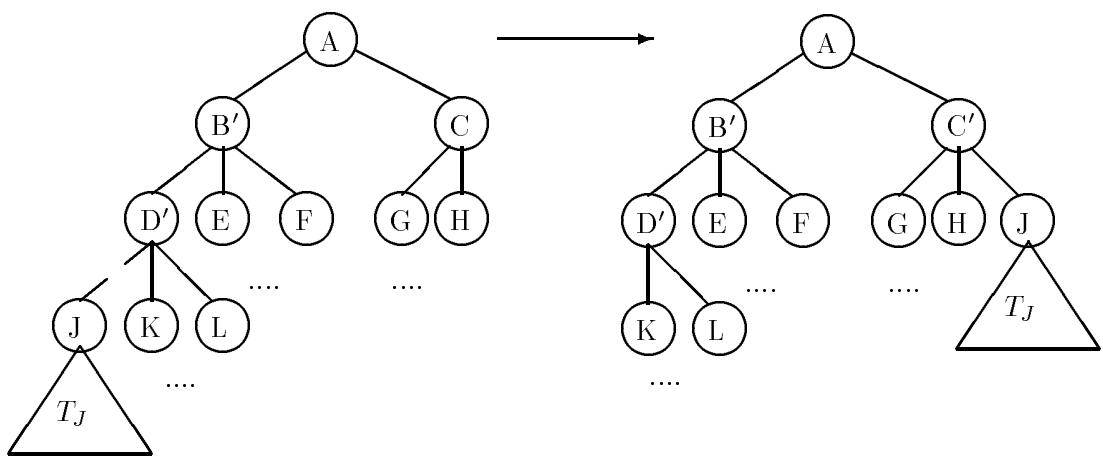

Figure 3: Hierarchical redistribution: the left subfigure indicates that cluster $J$ has just been removed as a descendent of $D$ and $B$, thus producing $D'$ and $B'$, and is about to be resorted relative to the children of the root ($A$). The rightmost figure shows $J$ has been placed as a new child of $C$. From Fisher (1995). Figure reproduced with permission from Proceedings of the First International Conference on Knowledge Discovery in Data Mining, Copyright ©1995 American Association for Artificial Intelligence.

The ISODATA algorithm (Duda & Hart, 1973) determines a target cluster for each observation, but does not actually change the clustering until targets for all observations have been determined; at this point, all observations are moved to their targets, thus altering the clustering. We limit ourselves to a sequential version, also described by Duda and Hart (1973), that moves each observation as its target is identified through sorting.

This strategy is conceptually simple, but is limited in its ability to overcome local maxima – the reclassification of a particular observation may be in the true direction of a better clustering, but it may not be perceived as such when the objective function is applied to the clustering that results from resorting the single observation.

### 3.3 Iterative Hierarchical Redistribution

An iterative optimization strategy that appears novel in the clustering literature is *iterative hierarchical redistribution*. This strategy is rationalized relative to single-observation iterative redistribution: even though moving a *set* of observations from one cluster to another may lead to a better clustering, the movement of any single observation may initially reduce clustering quality, thus preventing the eventual discovery of the better clustering. In response, hierarchical redistribution considers the movement of observation sets, represented by existing clusters in a hierarchical clustering.

Given an existing hierarchical clustering, a recursive loop examines sibling clusters in the hierarchy in a depth-first fashion. For each set of siblings, an inner, iterative loop examines each sibling, removes it from its current place in the hierarchy (along with its subtree), and resorts the cluster relative to the entire hierarchy. Removal requires that the various counts of ancestor clusters be decremented. Sorting the removed cluster is done based on the cluster's probabilistic description, and requires a minor generalization of the procedure for sorting individual observations: rather than incrementing certain variable value counts by 1 at a cluster to reflect the addition of a new observation, a 'host' cluster's variable value counts are incremented by the corresponding counts of the cluster being classified. A cluster may return to its original place in the hierarchy, or as Figure 3 illustrates, it may be sorted to an entirely different location.

The inner loop reclassifies each sibling of a set, and repeats until two consecutive iterations lead to the same set of siblings. The recursive loop then turns its attention to the children of each of these remaining siblings. Eventually, the individual observations represented by leaves are resorted (relative to the entire hierarchy) until there are no changes from one iteration to the next. Finally, the recursive loop may be applied to the hierarchy several times, thus defining an outermost (iterative) loop that terminates when no changes occur from one pass to the next.

There is one modification to this basic strategy that was implemented for reasons of cost: if there is no change in a subtree during a pass of the outermost loop through the hierarchy, then subsequent passes do not attempt to redistribute any clusters in this subtree unless and until a cluster (from some other location in the hierarchy) is placed in the subtree, thus changing the subtree's structure. In addition, there are cases where the $PU$ scores obtained by placing a cluster, $C$ (typically a singleton cluster), in either of two hosts will be the same. In such cases, the algorithm prefers placement of $C$ in its original host if this is one of the candidates with the high $PU$ score. This policy avoids infinite loops stemming from ties in the $PU$ score.

In sum, hierarchical redistribution takes large steps in the search for a better clustering. Similar to macro-operator learners (Iba, 1989) in problem-solving contexts, moving an observation set or cluster bridges distant points in the clustering space, so that a desirable change can be made that would not otherwise have been viewed as desirable if redistribution was limited to movement of individual observations. The redistribution of increasingly smaller, more granular clusters (terminating with individual observations) serves to increasingly refine the clustering.

To a large extent hierarchical redistribution was inspired by Fisher's (1987a) COBWEB system, which is fundamentally a hierarchical-sort-based strategy. However, COBWEB is augmented by operators of *merging, splitting*, and *promotion*. Merging combines two sibling clusters in a hierarchical clustering if to do so increases the quality of the partition of which the clusters are members; splitting can remove a cluster and promote its children to the next higher partition; a distinct promotion operator can promote an individual cluster to the next higher level. In fact, these could be regarded as 'iterative optimization' operators, but in keeping with COBWEB's cognitive modeling motivations, the cost of applying them is 'amortized' over time: as many observations are sorted, a cluster may migrate from one part of the hierarchical clustering to another through the collective and repeated application of merging, splitting, and promotion. A similar view is expressed by McKusick and Langley (1991), whose ARACHNE system differs from COBWEB, in part, by the way that it exploits the promotion operator. Unfortunately, in COBWEB, and to a lesser extent in ARACHNE, merging, splitting, and promotion are applied locally and migration through the hierarchy is limited in practice. In contrast, hierarchical redistribution resorts each cluster, regardless of its initial location in the tree, through the root of the entire tree, thus more vigorously pursuing migration and more globally evaluating the merits of such moves.[1]

The idea of hierarchical redistribution is also closely related to strategies found in the BRIDGER (Reich & Fenves, 1991) and HIERARCH (Nevins, 1995) systems. In particular, BRIDGER identifies 'misplaced' clusters in a hierarchical clustering using a criterion specified, in part, by a domain expert, whereas hierarchical redistribution simply uses the objective function. In BRIDGER each misplaced cluster is removed (together with its subtree), but the cluster/subtree is not resorted as a single unit; rather, the observations covered by the cluster are resorted individually. This approach captures, in part, the idea of hierarchical redistribution, though the resorting of individual observations may not escape local optima to the same extent as hierarchical redistribution.

Given an existing hierarchical clustering and a new observation, HIERARCH conducts a branch-and-bound search through the clustering, looking for the cluster that 'best matches' the observation. When the best host is found, clusters in the 'vicinity' of this best host are reclassified using branch-and-bound with respect to the entire hierarchy. These clusters need not be singletons, and their reclassification can spawn other reclassifications until a termination condition is reached.

It is unclear how HIERARCH's procedure scales up to large data sets; the number of experimental trials and the size of test data sets is considerably less than we describe shortly. Nonetheless, the importance of bridging distant regions of the clustering space by reclassifying observation sets *en masse* in made explicit. Like COBWEB, HIERARCH is incremental, changes to a hierarchy are triggered along the path that classifies a new observation, and these changes may move many observations simultaneously, thus 'amortizing' the cost of optimization over time. In contrast, hierarchical redistribution is motivated by a philosophy that sorting (or some other method) can produce a tentative clustering over all the data quickly, followed by iterative optimization procedures in background that revise the clustering intermittently. While hierarchical redistribution reflects many of the same ideas implemented in HIERARCH, COBWEB, and related systems, it appears novel as an iterative optimization strategy that is decoupled from any particular initial clustering strategy.

### 3.4 Comparisons between Iterative Optimization Strategies

This section compares iterative optimization strategies under two experimental conditions. In the first condition, a random ordering of observations is generated and hierarchically sorted. Each of the optimization strategies is then applied independently to the resultant hierarchical clustering. These experiments assume that the primary goal of clustering is to discover a single-level partitioning of the data that is of optimal quality. Thus, the *objective function score of the first-level partition* is taken as the most important dependent variable. An independent variable is the height of the initially-constructed clustering; this can effect the granularity of clusters that are used in hierarchical redistribution. A hierarchical clus-tering of height 2 corresponds to a single level partition of the data at depth 1 (the root is at depth 0), with leaves corresponding to individual observations at depth 2.

---

1. Considering global changes also motivated redistribution of individual observations in ITERATE. As Nevins (1995) notes in commentary on experimental comparisons between of ITERATE and COBWEB (Fisher et al., 1992), even global movement of single observations typically did not perform as well as local movement of sets of observations simultaneously, as implemented by COBWEB's merging and splitting operators.

|  |  | Random | Similarity |
|---|---|---|---|
| Soybean (small) (47 obs, 36 vars) | sort | 1.53 (0.11) | 1.08 (0.18) |
|  | reorder/resort | 1.61 (0.02) | 1.56 (0.08) |
|  | iter. redist. | 1.54 (0.10) | 1.34 (0.20) |
|  | hier. redist. | 1.60 (0.05) | 1.50 (0.08) |
| Soybean (large) (307 obs, 36 vars) | sort | 0.89 (0.08) | 0.66 (0.14) |
|  | reorder/resort | 0.97 (0.04) | 0.96 (0.05) |
|  | iter. redist. | 0.92 (0.07) | 0.84 (0.10) |
|  | hier. redist. | 1.06 (0.02) | 1.06 (0.01) |
| House (435 obs, 17 vars) | sort | 1.22 (0.30) | 0.83 (0.16) |
|  | reorder/resort | 1.66 (0.09) | 1.57 (0.18) |
|  | iter. redist. | 1.24 (0.28) | 1.06 (0.19) |
|  | hier. redist. | 1.68 (0.00) | 1.68 (0.00) |
| Mushroom (1000 obs, 23 vars) | sort | 1.10 (0.13) | 0.73 (0.22) |
|  | reorder/resort | 1.10 (0.08) | 1.16 (0.08) |
|  | iter. redist. | 1.10 (0.12) | 0.95 (0.19) |
|  | hier. redist. | 1.27 (0.00) | 1.24 (0.10) |

Table 2: Iterative optimization strategies with initial clusterings generated from sorting random and similarity ordered observations. Tree height is 2. Averages and standard deviations of $PU$ scores over 20 trials.

In addition to experiments on clusterings derived by sorting random initial orderings, each redistribution strategy was tested on exceptionally poor initial clusterings generated by nonrandom orderings. Just as 'dissimilarity' orderings lead to good clusterings, 'similarity' orderings lead to poor clusterings (Fisher et al., 1992). Intuitively, a similarity ordering samples observations within the same region of the data description space before sampling observations from differing regions. The reordering procedure of Section 3.1 is easily modified to produce similarity orderings by ranking each set of siblings in a hierarchical clustering from least to most probable, and appending rather than interleaving observation lists from differing clusters as the algorithm pops up the recursive levels. A similarity ordering is produced by applying this procedure to an initial clustering produced by an earlier sort of a random ordering. Another clustering is then produced by sorting the similarity-ordered data, and the three iterative optimization strategies are applied independently. We do not advocate that one build clusterings from similarity orderings in practice, but experiments with such orderings better test the robustness of the various optimization strategies.

Table 2 shows the results of experiments with random and similarity orderings of data from four databases of the UCI repository.[2] These results assume an initial clustering of height 2 (i.e., a top-level partition + observations at leaves). Each cell represents an average and standard deviation over 20 trials. The first cell (labeled 'sort') of each domain is the mean $PU$ scores initially obtained by sorting. Subsequent rows under each domain reflect the mean scores obtained by the reordering/resorting procedure of Section 3.1, iterative redistribution of single observations described in Section 3.2, and hierarchical redistribution described in Section 3.3.

---

2. A reduced `mushroom` data set was obtained by randomly selecting 1000 observations from the original data set.

The main findings reflected in Table 2 are:

1. Initial hierarchical sorting from random input does reasonably well; $PU$ scores in this case are closer to the scores of optimized trees, than to the poorest scores obtained after sorting on similarity orderings. This weakly suggests that initial sorting on random input takes a substantial step in the space of clusterings towards discovery of the final structure.

2. Hierarchical redistribution achieves the highest mean $PU$ score in both the random and similarity case in 3 of the 4 domains. The small soybean domain is the exception.

3. In the House domain (random and similarity case) and the Mushroom domain (random case only), the standard deviation in $PU$ scores of clusterings optimized by hierarchical redistribution is 0.00, indicating that it has always constructed level-1 partitions of the same $PU$ score in all 20 trials.

4. Reordering and reclustering comes closest to hierarchical redistribution's performance in all cases, bettering it in the Small Soybean domain.

5. Single-observation redistribution modestly improves an initial sort, and is substantially worse than the other two optimization methods.

Note that with initial hierarchical clusterings of height 2, the only difference between iterative hierarchical redistribution and redistribution of single observations is that hierarchical redistribution considers 'merging' clusters of the partition (by reclassifying one with respect to the others) prior to redistributing single observations during each pass through the hierarchy.

Section 3.3 suggested that the expected benefits of hierarchical redistribution might be greater for deeper initial trees with more granular clusters. Table 3 shows results on the same domains and initial orderings when tree height is 4 for hierarchical redistribution; for the reader's convenience we also repeat the results from Table 2 for hierarchical redistribution when tree height is 2. In moving from height 2 to 4, there is modest improvement in the small Soybean domain (particularly under Similarity orderings), and very slight improvement in the large Soybean domain and the Mushroom domain under Similarity orderings.[3] While the improvements are very modest, moving to height 4 trees leads to near identical performance in the random and similarity ordering conditions. This suggests that hierarchical redistribution is able to effectively overcome the disadvantage of initially poor clusterings.

Experiments with reorder/resort and iterative distribution of single observations also were varied with respect to tree height (e.g., height 3). For each of these methods, the deepest set of clusters in the initial hierarchy above the leaves, was taken as the initial partition. Reordering/resorting scores remained roughly the same as the height 2 condition, but clusterings produced by single-observation redistribution had *PU* scores that were considerably worse than those given in Table 2.

---

3. A standard deviation of $0.0\overline{0}$ indicates that the standard deviation was non-0, but not observable at the 2nd decimal place after rounding.

|  | Random | | Similarity | |
|---|---|---|---|---|
|  | height 2 | height 4 | height 2 | height 4 |
| Soybean (small) | 1.60 (0.05) | 1.62 (0.00) | 1.50 (0.08) | 1.62 (0.00) |
| Soybean (large) | 1.06 (0.02) | 1.07 (0.02) | 1.06 (0.01) | 1.07 (0.01) |
| House | 1.68 (0.00) | 1.68 (0.00) | 1.68 (0.00) | 1.68 (0.00) |
| Mushroom | 1.27 (0.00) | 1.27 (0.00) | 1.24 (0.10) | 1.27 (0.00) |

Table 3: Hierarchical redistribution with initial clusterings generated from sorting random and similarity ordered observations. Results are shown for tree heights of 2 (copied from Table 2) and 4. Averages and standard deviations of *PU* scores over 20 trials.

We also recorded execution time for each method. Table 4 shows the time required for each method in seconds.[4] In particular, for each domain, Table 4 lists the mean time for initial sorting, and the mean additional time for each optimization method. Ironically, these experiments demonstrate that even though hierarchical redistribution 'bottoms-out' in a single-observation form of redistribution, the former is consistently faster than the latter for trees of height 2 – reclassifying a cluster simultaneously moves a set of observations, which would otherwise have to be repeatedly evaluated for redistribution individually with increased time to stabilization.[5]

Table 4 assumes the tree constructed by initial sorting is bounded to height 2. Table 5 gives the time requirements of hierarchical sorting and hierarchical redistribution when the initial tree is bounded to height 4. As the tree gets deeper the cost of hierarchical redistribution grows substantially, and as our comparison of performance with height 2 and 4 trees in Table 3 suggests, there are drastically diminishing returns in terms of partition quality. Importantly, limited experiments with trees of height 2, 3, and 4 indicate that the cost of hierarchical redistribution is comparable to the cost of reorder/resort at greater tree heights and significantly less expensive than single-observation redistribution. It is difficult to give a cost analysis of hierarchical redistribution (and the other methods for that matter), since bounds on loop iterations probably depend on the nature of the objective function. Suffice it to say that the number of nodes that are subject to hierarchical redistribution in a tree covering $n$ observations is bounded above by $2n - 1$; there may be up to $n$ leaves and up to $n - 1$ internal nodes given that each internal node has no less than 2 children.

If iterative optimization is to occur in background, real-time response is not important, and cluster quality is paramount, then it is probably worth applying hierarchical redis-tribution to deeper trees; this is consistent with the philosophy behind such systems as AUTOCLASS and SNOB. For the domains examined here, however, it does not seem cost effective to optimize with trees of height greater than 4. Thus, we adopt a tree construction strategy that builds a hierarchical clustering three levels at a time (with hierarchical redistribution) in the experiments of Section 4.

---

4. Routines were implemented in SUN Common Lisp, compiled, and run on a SUN 3/60.
5. Similar timing results occur in other computational contexts as well. Consider the relation between insertion sort and Shell sort. Shell sort's final 'pass' of a table is an insertion sort that is limited to moving table elements between consecutive table locations at a time. The large efficiency advantage of Shell Sort stems from the fact that previous passes of the table have moved elements large distances, thus by the final pass, the table is nearly sorted.

|  |  | Random | Similarity |
|---|---|---|---|
| Soybean (small) (47 obs, 36 vars) | sort | 6.98 (1.43) | 7.21 (1.31) |
|  | reorder/resort | 14.82 (2.60) | 18.27 (6.00) |
|  | iter. redist. | 9.00 (5.94) | 15.51 (7.72) |
|  | hier. redist. | 6.99 (1.28) | 8.87 (3.58) |
| Soybean (large) (307 obs, 36 vars) | sort | 50.62 (6.11) | 54.09 (13.25) |
|  | reorder/resort | 141.36 (46.99) | 153.22 (43.59) |
|  | iter. redist. | 166.53 (55.53) | 307.59 (160.66) |
|  | hier. redist. | 79.00 (19.23) | 87.27 (19.64) |
| House (435 obs, 17 vars) | sort | 34.99 (7.55) | 39.15 (7.60) |
|  | reorder/resort | 87.78 (23.94) | 97.63 (29.54) |
|  | iter. redist. | 177.75 (94.53) | 320.43 (124.78) |
|  | hier. redist. | 55.90 (11.92) | 73.54 (10.05) |
| Mushroom (1000 obs, 23 vars) | sort | 111.47 (19.19) | 119.33 (25.86) |
|  | reorder/resort | 301.34 (100.56) | 391.80 (211.54) |
|  | iter. redist. | 162.58 (85.20) | 390.11 (191.62) |
|  | hier. redist. | 91.87 (29.50) | 151.45 (48.89) |

Table 4: Time requirements (in seconds) of hierarchical sorting and iterative optimization with initial clusterings generated from sorting random and similarity ordered observations. Tree height is 2. Averages and standard deviations over 20 trials.

### 3.5 Discussion of Iterative Optimization Methods

Our experiments demonstrate the relative abilities of three iterative optimization strategies, which have been coupled with the *PU* objective function and hierarchical sorting to generate initial clusterings. The reorder/resort optimization strategy of Section 3.1 makes most sense with sorting as the primary clustering strategy, but the other optimization techniques are not strongly tied to a particular initial clustering strategy. For example, hierarchical redistribution can also be applied to hierarchical clusterings generated by an agglomerative strategy (Duda & Hart, 1973; Everitt, 1981; Fisher et al., 1992), which uses a bottom-up procedure to construct hierarchical clusterings by repeatedly 'merging' observations and resulting clusters until an all-inclusive root cluster is generated. Agglomerative methods do not suffer from ordering effects, but they are greedy algorithms, which are susceptible to the limitations of local decision making generally, and would thus likely benefit from iterative optimization.

|  |  | Random | | Similarity | |
|---|---|---|---|---|---|
|  |  | height 2 | height 4 | height 2 | height 4 |
| Soy (small) | sort | 6.98 (1.4) | 18 (2) | 7.21 (1.3) | 21 (2) |
|  | hier. redist. | 6.99 (1.3) | 94 (28) | 8.87 (3.6) | 133 (28) |
| Soy (large) | sort | 50.62 (6.1) | 142 (10) | 54.09 (13.3) | 152 (11) |
|  | hier. redist. | 79.00 (19.2) | 436 (139) | 87.27 (19.6) | 576 (260) |
| House | sort | 34.99 (7.6) | 104 (9) | 39.15 (7.6) | 120 (12) |
|  | hier. redist. | 55.90 (11.9) | 355 (71) | 73.54 (10.1) | 425 (105) |
| Mushroom | sort | 111.47 (19.2) | 407 (64) | 119.33 (25.9) | 443 (65) |
|  | hier. redist. | 91.87 (29.5) | 1288 (458) | 151.45 (48.9) | 1368 (335) |

Table 5: Time requirements (in seconds) of hierarchical sorting and hierarchical redistribution with initial clusterings generated from sorting random and similarity ordered observations. Results are shown for tree heights of 2 (copied from Table 4) and 4. Averages and standard deviations over 20 trials.

In addition, all three optimization strategies can be applied regardless of objective function. Nonetheless, the relative benefits of these methods undoubtedly varies with objective function. For example, the *PU* function has the undesirable characteristic that it may, under very particular circumstances, view two partitions that are very close in form as separated by a 'cliff' (Fisher, 1987b; Fisher et al., 1992). Consider a partition of $M$ observations involving only two, roughly equal-sized clusters; its *PU* score has the form $PU(\{C_1, C_2\}) = [\sum_{k=1}^{2} CU(C_k)]/2$. If we create a partition of three clusters by removing a single observation from, say $C_2$, and creating a new singleton cluster, $C_3$ we have $PU(\{C_1, C_2', C_3\}) = [\sum_{k=1}^{3} CU(C_k)]/3$. If $M$ is relatively large, $CU(C_3)$ will have a very small score due to the term, $P(C_3) = 1/M$ (see Section 2.1). Because we are taking the average $CU$ score of clusters, the difference between $PU(\{C_1, C_2\})$ and $PU(\{C_1, C_2', C_3\})$ may be quite large, even though they differ in the placement of only one observation. Thus, limiting experiments to the *PU* function may exaggerate the general advantage of hierarchical redistribution relative to the other two optimization methods. This statement is simultaneously a positive statement about the robustness of hierarchical redistribution in the face of an objective function with cliffs, and a negative statement about *PU* for defining such discontinuities. Nonetheless, *PU* and variants have been adopted in systems that fall within the COBWEB family (Gennari et al., 1989; McKusick & Thompson, 1990; Reich & Fenves, 1991; Iba & Gennari, 1991; McKusick & Langley, 1991; Kilander, 1994; Ketterlin et al., 1995; Biswas et al., 1994). Section 5.2 suggests some alternative objective functions.

Beyond the nonoptimality of *PU*, our findings should not be taken as the best that these strategies can do when they are engineered for a particular clustering system. We could introduce forms of randomization or systematic variation to any of the three strategies. For example, while Michalski and Stepp's seed-selection methodology inspires reordering/resorting, Michalski and Stepp's approach selects 'border' observations when the selection of 'centroids' fails to improve clustering quality from one iteration to the next;

this is an example of the kind of systematic variations that one might introduce in pursuit of better clusterings. In contrast, AUTOCLASS may take large heuristically-guided 'jumps' away from a current clustering. This approach might be, in fact, a somewhat less systematic (but equally successful) variation on the macro-operator theme that inspired hierarchical redistribution, and is similar to HIERARCH's approach as well. SNOB (Wallace & Dowe, 1994) employs a variety of search operators, including operators similar to COBWEB's merge and split (though without the same restrictions on local application), random restart of the clustering process with new seed observations, and 'redistribution' of observations.[6] In fact, the user can program SNOB's search strategy using these differing primitive search operators. In any case, systems such as CLUSTER/2, AUTOCLASS, and SNOB do not simply 'give up' when they fail to improve clustering quality from one iteration to the next.

As SNOB illustrates, one or more strategies might be combined to advantage. As an additional example, Biswas et al. (1994) adapt Fisher, Xu, and Zard's (1992) dissimilarity ordering strategy to preorder observations prior to clustering. After sorting using $PU$, their ITERATE system then applies iterative redistribution of single observations using a *category match* measure by Fisher and Langley (1990).

The combination of preordering and iterative redistribution appears to yield good results in ITERATE. Our results with reorder/resort suggest that preordering is primarily responsible for quality benefits over a simple sort, but the relative contribution of ITERATE's redistribution operator is not certain since it differs in some respects from the redistribution technique described in this paper.[7] However, the use of three different measures – distance, $PU$, and category match – during clustering may be unnecessary and adds undesirable coupling in the design of the clustering algorithm. If, for example, one wants to experiment with the merits of differing objective functions, it is undesirable to worry about the 'compatibility' of this function with two other measures. In contrast, reordering/resorting generalizes Fisher et al.'s (1992) ordering strategy; this generalization and the iterative redistribution strategy we describe assume no auxiliary measures beyond the objective function. In fact, as in Fisher (1987a, 1987b), an evaluation of ITERATE's clusterings is made using measures of variable value predictability or $P(A_i = V_{ij}|C_k)$, predictiveness or $P(C_k|A_i = V_{ij})$, and their product. It is not clear that a system need exploit several related, albeit different measures during the generation and evaluation of clusterings; undoubtedly a single, carefully selected objective function can be used exclusively during clustering.

Reordering/resorting and iterative redistribution of single observations could be combined in a manner similar to ITERATE's exploitation of certain specializations of these procedures. Our results suggest that reordering/resorting would put a clustering in a good 'ballpark', while iterative redistribution would subsequently make modest refinements. We have not combined strategies, but in a sense conducted the inverse of an 'ablation' study, by evaluating individual strategies in isolation. In the limited number of domains explored in Section 3.4, however, it appears difficult to better hierarchical redistribution.

Finally, our experiments applied various optimization techniques after all data was sorted. It may be desirable to apply the optimization procedures at intermittent points during sorting. This may improve the quality of final clusterings using reordering/resorting and redistribution of single observations, as well as reduce the overall cost of constructing final optimized clusterings using any of the methods, including hierarchical redistribution, which already appears to do quite well on the quality dimension. In fact, HIERARCH can be viewed as performing something akin to a restricted form of hierarchical redistribution after each observation. This is probably too extreme – if iterative optimization is performed too often, the resultant cost can outweigh any savings gleaned by maintaining relatively well optimized trees throughout the sorting process. Utgoff (1994) makes a similar suggestion for intermittent restructuring of decision trees during incremental, supervised induction.

---

6. Importantly, SNOB (and AUTOCLASS) assumes probabilistic assignment of observations to clusters.
7. ITERATE uses a measure for redistribution (Fisher & Langley, 1990) that probably smoothes 'cliffs', and it uses an ISODATA, non-sequential version of redistribution.

## 4. Simplifying Hierarchical Clusterings

A hierarchical clustering can be grown to arbitrary height. If there is structure in the data, then ideally the top layers of the clustering reflect this structure (and substructure as one descends the hierarchy). However, lower levels of the clustering may not reflect meaningful structure. This is the result of overfitting, which one finds in supervised induction as well. Inspired by certain forms of retrospective (or post-tree-construction) pruning in decision-tree induction, we use resampling to identify 'frontiers' of a hierarchical clustering that are good candidates for pruning. Following initial hierarchy construction and iterative optimization, this simplification process is a final phase of search through the space of hierarchical clusterings intended to ease the burden of a data analyst.

### 4.1 Identifying Variable Frontiers by Resampling

Several authors (Fisher, 1987a; Cheeseman et al., 1988; Anderson & Matessa, 1991) motivate clustering as a means of improving performance on a task akin to pattern completion, where the error rate over completed patterns can be used to 'externally' judge the utility of a clustering. Given a probablistic categorization tree of the type we have assumed, a new observation with an unknown value for a variable can be classified down the hierarchy using a small variation on the hierarchical sorting procedure described earlier.[8] Classification is terminated at a selected node (cluster) along the classification path, and the variable value of highest probability at that cluster is predicted as the unknown variable value of the new observation. Naively, classification might always terminate at a leaf (i.e., an observation), and the leaf's value along the specified variable would be predicted as the variable value of the new observation. Our use of a simple resampling strategy known as *holdout* (Weiss & Kulikowski, 1991) is motivated by the fact that a variable might be better predicted at some internal node in the classification path. The identification of ideal-prediction frontiers for each variable suggests a pruning strategy for hierarchical clusterings.

Given a hierarchical clustering and a *validation* set of observations, the validation set is used to identify an appropriate *frontier* of clusters for prediction of each variable. Figure 4 illustrates that the preferred frontiers of any two variables may differ, and clusters within a frontier may be at different depths. For *each* variable, $A_i$, the objects from the validation set are each classified through the hierarchical clustering with the value of variable $A_i$ 'masked' for purposes of classification; at each cluster encountered during classification the observation's value for $A_i$ is compared to the most probable value for $A_i$ at the cluster; if they are the same, then the observation's value would have been correctly predicted at the cluster. A count of all such correct predictions for each variable at a cluster is maintained. Following classification for all variables over all observations of the validation set, a preferred frontier for each variable is identified that maximizes the number of correct counts for the variable. This is a simple, bottom-up procedure that insures that the number of correct counts at a node on the variable's frontier is greater than or equal to the *sum* of correct counts for the variable over each set of mutually-exclusive, collectively-exhaustive descendents of the node.

---

8. Classification is identical to sorting except that the observation is not added to the clustering and statistics at each node encountered during sorting are not permanently updated to reflect the new observation.

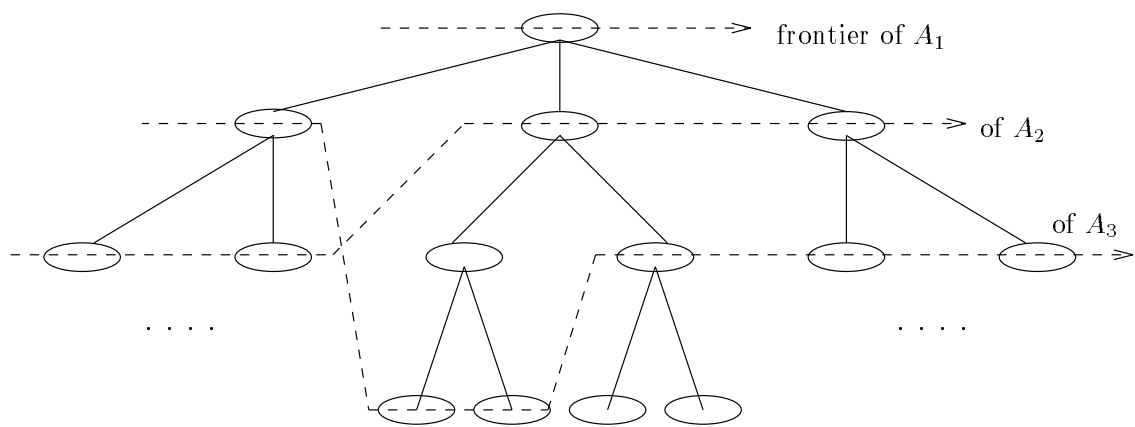

Figure 4: Frontiers for three variables in a hypothetical clustering. From Fisher (1995). Figure reproduced with permission from Proceedings of the First International Conference on Knowledge Discovery in Data Mining, Copyright ©1995 American Association for Artificial Intelligence.

Variable-specific frontiers enable a number of pruning strategies. For example, a node that lies below the frontier of every variable offers no apparent advantage in terms of pattern-completion error rate; such a node probably reflects no meaningful structure and it (and its descendents) may be pruned. However, if an analyst is focusing attention on a subset of the variables, then frontiers might be more flexibly exploited for pruning.

### 4.2 Experiments with Validation

To test the validation procedure's promise for simplifying hierarchical clusterings, each of the data sets used in the optimization experiments of Section 3.4 was randomly divided into three subsets: 40% for training, 40% for validation, and 20% for test. A hierarchical clustering is first constructed by sorting the training set in randomized order. This hierarchy is then optimized using iterative hierarchical redistribution. Actually, because of cost considerations, a hierarchy is constructed several levels at a time. The hierarchy is initially constructed to height 4, where the deepest level is the set of training observations. This hierarchy is optimized using hierarchical redistribution. Clusters at the bottommost level (i.e., 4) are removed as children of level 3 clusters, and the subset of training observations covered by each cluster of level 3 is hierarchically sorted to a height 4 tree and optimized. The roots of these subordinate clusterings are then substituted for each cluster at depth 3 in the original tree. The process is repeated on clusters at level 3 of the subordinate trees and subsequent trees thereafter until no further decomposition is possible. The final hierarchy, which is not of constant-bounded height, decomposes the entire training set to singleton clusters, each containing a single training observation. The validation set is then used to identify variable frontiers within the entire hierarchy.

|  | Unvalidated | Validated |
|---|---|---|
| Soybean (small) | | |
| Leaves | 18.00 (0.00) | 13.10 (1.59) |
| Accuracy | 0.85 (0.01) | 0.85 (0.01) |
| Ave. Frontier Size | 18.00 (0.00) | 2.75 (1.17) |
| Soybean (large) | | |
| Leaves | 122.00 (0.00) | 79.10 (5.80) |
| Accuracy | 0.83 (0.02) | 0.83 (0.02) |
| Ave. Frontier Size | 122.00 (0.00) | 17.01 (4.75) |
| House | | |
| Leaves | 174.00 (0.00) | 49.10 (7.18) |
| Accuracy | 0.76 (0.02) | 0.81 (0.01) |
| Ave. Frontier Size | 174.00 (0.00) | 9.90 (5.16) |
| Mushroom | | |
| Leaves | 400.00 (0.00) | 96.30 (11.79) |
| Accuracy | 0.80 (0.01) | 0.82 (0.01) |
| Ave. Frontier Size | 400.00 (0.00) | 11.07 (4.28) |

Table 6: Characteristics of optimized clusterings before and after validation. Average and standard deviations over 20 trials.

During testing of a validated clustering, each variable of each test observation is masked in turn; when classification reaches a cluster on the frontier of the masked variable, the most probable value is predicted as the value of the observation; the proportion of correct predictions for each variable over the test set is recorded. For comparative purposes, we also use the test set to evaluate predictions stemming from the unvalidated tree, where all variable predictions are made at the leaves (singleton clusters) of this tree.

Table 6 shows results from 20 experimental trials using optimized, unvalidated and validated clusterings generated as just described from random orderings. The first row of each domain lists the average number of leaves (over the 20 experimental trials) for the unvalidated and validated trees. The unvalidated clusterings decompose the training data to single-observation leaves – the number of leaves equals the number of training observations. In the validated clustering, we assume that clusters are pruned if they lie below the frontiers of *all* variables. Thus, a leaf in a validated clustering is a cluster (in the original clustering) that is on the frontier of *at least one* variable, and none of its descendent clusters (in the original clustering) are on the frontier of any variable. For example, if we assume that the tree of Figure 4 covers data described only in terms of variables $A_1$, $A_2$, and $A_3$, then the number of leaves in this validated clustering would be 7.

Prediction accuracies in the second row of each domain entry are the mean proportion of correct predictions over *all* variables over 20 trials. Predictions were generated at leaves (singleton clusters) in the unvalidated hierarchical clusterings and at appropriate variable frontiers in the validated clusterings. In all cases, validation/pruning substantially reduces clustering size and it does not diminish accuracy.

The number of leaves in the validated case, as we have described it, assumes a very coarse pruning strategy; it will not necessarily discriminate a clustering with uniformly deep frontiers from one with a single or very few deep frontiers. We have suggested that more flexible pruning or 'attention' strategies might be possible when an analyst is focusing on one or a few variables. We will not specify such strategies, but the statistic given in row 3 of each domain entry suggests that clusterings can be rendered in considerably simpler forms when an analyst's attention is selective. Row 3 is the *average number of frontier clusters per variable*. This is an average over all variables and all experimental trials.[9] In the validated tree of Figure 4 the average frontier size is $(1 + 4 + 6)/3 = 3.67$.

Intuitively, a frontier cluster of a variable is a 'leaf' as far as prediction of that variable is concerned. The 'frontier size' for unvalidated clusterings is simply given by the number of leaves, since this is where all variable predictions are made in the unvalidated case. Our results suggest that when attention is selective, a partial clustering that captures the structure involving selected variables can be presented to an analyst in very simplified form.

### 4.3 Discussion of Validation

The resampling-based validation method is inspired by earlier work by Fisher (1989), which identified variable frontiers within a strict incremental (i.e., sorting) context – no separate validation set was reserved, but rather the training set was used for identifying variable frontiers as well. In particular, as each training observation was hierarchically sorted using COBWEB, each of the observation's variable values were predicted and 'correct' counts at each node were updated for all correctly anticipated variables. In Fisher (1989) variable values were not masked during sorting – knowledge of each variable value was used during sorting, thus helping to guide classification, and validation. In addition, the hierarchy changed during sorting/validation. While this incremental strategy led to desirable results in terms of pattern-completion error rate, it is likely that the variable frontiers identified by the incremental method are less desirable than frontiers identified with holdout, where we strictly segregate the training and validation sets of observations. In addition to Fisher (1989), our work on variable frontiers can be traced back to ideas by Lebowitz (1982) and Kolodner (1983), and more directly to Fisher (1987b), Fisher and Schlimmer (1988), and Reich and Fenves (1991), each of which use a very different method to identify something similar in spirit to frontiers as defined here.

Our method of validation and pruning is inspired by retrospective pruning strategies in decision tree induction such as *reduced error pruning* (Quinlan, 1987, 1993; Mingers, 1989a). In a Bayesian clustering system such as AUTOCLASS (Cheeseman et al., 1988), or the minimum message length (MML) approach adopted by Snob (Wallace & Dowe, 1994), the expansion of a hierarchical clustering is mediated by a tradeoff between prior belief in the existence of further structure and evidence in the data for further structure. We will not detail this fundamental tradeoff, but suffice it to say that expansion of a hierarchical clustering will cease along a path when the evidence for further structure in the data is insufficient in the face of prior bias. Undoubtedly, the Bayesian and MML approaches can be adapted to identify variable-specific frontiers, and thus be used in the kind of flexible pruning and focusing strategies that we have implied. In fact, something very similar in intent has been implemented in Autoclass (Hanson, Stutz, & Cheeseman, 1991) as a way of reducing the cost of clustering with this system: variables that covary may be 'blocked', or in some sense treated as one. This version of Autoclass searches a space of hierarchical clusterings, with blocks of variables assigned to particular clusters in the hierarchy. The interpretation of such assignments is that a cluster 'inherits' the variable value distributions of variable blocks assigned to the cluster's ancestors. Inversely, the basic idea is that one need not proceed below a cluster to determine the value distributions of variables assigned to that cluster.

---

9. The 'standard deviations' given in Row 3 are actually the mean of the standard deviations over the frontier sizes for individual variables.

Our experimental results suggest the utility of resampling for validation, the identification of variable frontiers, and pruning. However, the procedure described is not a method *per se* of clustering over all the available data, since it requires that a validation set be held out during initial hierarchy construction.[10] There are several options that seem worthy of experimental evaluation in adapting this validation strategy as a tool for simplification of hierarchical clusterings. One strategy would be to hold out a validation set, cluster over a training set, identify variable frontiers with the validation set, and then sort the validation set relative to the clustering. This single holdout methodology has its problems, however, for reasons similar to those identified for single holdout in supervised settings (Weiss & Kulikowski, 1991).

A better strategy might be one akin to *n-fold-cross-validation*: a hierarchical clustering is constructed over all available data, then each observation is removed,[11] it is used for validation with respect to each variable, and then the observation is reinstated in its original location (together with the original variable value statistics of clusters along the path to this location).

## 5. General Discussion

The evaluation of the various strategies discussed in this paper reflect two paradigms for validating clusterings. *Internal* validation is concerned with evaluating the merits of the control strategy that searches the space of clusterings: evaluating the extent that the search strategy uncovers clusterings of high quality as measured by the objective function. Internal validation was the focus of Section 3.4. *External* validation is concerned with determining the utility of a discovered clustering relative to some performance task. We have noted that several authors point to minimization of error rate in pattern completion as a generic performance task that motivates their choice of objective function. External validation was the focus of Section 4.2.

---

10. For purposes of evaluating the merits of our validation strategy in terms of error rate, we also held out a separate test set. Having demonstrated the point, however, we would not require that a separate test set be held out when using resampling as a validation strategy.
11. The observation is physically removed, and variable value statistics at clusters that lie along the path from root to the observation are decremented.

|  | Unoptimized | | Optimized | |
|---|---|---|---|---|
|  | Unvalidated | Validated | Unvalidated | Validated |
| Soybean (small) | | | | |
| Leaves | 18.00 (0.00) | 15.35 (1.81) | 18.00 (0.00) | 13.10 (1.59) |
| Accuracy | 0.84 (0.18) | 0.85 (0.01) | 0.85 (0.01) | 0.85 (0.01) |
| Ave. Frontier Size | 18.00 (0.00) | 3.97 (1.62) | 18.00 (0.00) | 2.75 (1.17) |
| Soybean (large) | | | | |
| Leaves | 122.00 (0.00) | 88.55 (4.46) | 122.00 (0.00) | 79.10 (5.80) |
| Accuracy | 0.82 (0.02) | 0.82 (0.02) | 0.83 (0.02) | 0.83 (0.02) |
| Ave. Frontier Size | 122.00 (0.00) | 24.74 (7.52) | 122.00 (0.00) | 17.01 (4.75) |
| House | | | | |
| Leaves | 174.00 (0.00) | 68.95 (8.15) | 174.00 (0.00) | 49.10 (7.18) |
| Accuracy | 0.76 (0.02) | 0.81 (0.02) | 0.76 (0.02) | 0.81 (0.01) |
| Ave. Frontier Size | 174.00 (0.00) | 17.72 (7.81) | 174.00 (0.00) | 9.90 (5.16) |
| Mushroom | | | | |
| Leaves | 400.00 (0.00) | 145.50 (20.64) | 400.00 (0.00) | 96.30 (11.79) |
| Accuracy | 0.80 (0.01) | 0.82 (0.01) | 0.80 (0.01) | 0.82 (0.01) |
| Ave. Frontier Size | 400.00 (0.00) | 22.85 (8.75) | 400.00 (0.00) | 11.07 (4.28) |

Table 7: Characteristics of unoptimized and optimized clusterings before and after validation. Average and standard deviations over 20 trials.

This section explores validation issues more closely, identifies both error rate and simplicity (or 'cost') as necessary external criteria for discriminating clustering utility, suggests a number of alternative objective functions that might be usefully compared using these criteria, and speculates that these external validation criteria (taken collectively) reflect reasonable criteria that data analysts may use to judge the utility of clusterings.

### 5.1 A Closer Look at External Validation Criteria

Ideally, clustering quality as measured by the objective function should be well correlated with clustering utility as determined by a performance task: the higher the quality of a clustering as judged by the objective function, the greater the performance improvement (e.g., reduction of error rate), and the lower the quality, the less that performance improves. However, several authors (Fisher et al., 1992; Nevins, 1995; Devaney & Ram, 1993) have pointed out that $PU$ scores do not seem well-correlated with error rates. More precisely, hierarchical clusterings (constructed by hierarchical sorting) in which the top-level partition has a low $PU$ score lead to roughly the same error rates as hierarchies in which the top-level partition has a high $PU$ score, when variable-value predictions are made at leaves (singleton clusters). Apparently, even with poor partitions at each level as measured by $PU$, test observations are classified to the same or similar observations at the leaves of a hierarchical clustering. Pattern-completion error rate under these circumstances seems insufficient to discriminate what we might otherwise consider to be good and poor clusterings.

Our work on simplification in Section 4 suggests that in addition to error rate, we might choose to judge competing hierarchical clusterings based on simplicity or some similarly-intended criterion. Both error rate and simplicity are used to judge classifiers in supervised contexts. We have seen that holdout can be used to substantially 'simplify' a hierarchical clustering. The question we now ask is whether hierarchical clusterings that have been optimized relative to $PU$ can be simplified more substantially than unoptimized clusterings with no degradation in pattern-completion error rate?

To answer this question we repeated the validation experiments of Section 4.2 under a second experimental condition: hierarchical clusterings were constructed from similarity orderings of the observations using hierarchical sorting. We saw in Section 3.4 that similarity orderings tend to result in clusterings judged poor by the $PU$ function. We do not optimize these hierarchies using hierarchical optimization. Table 7 shows accuracies, number of leaves, and average frontier sizes, for unoptimized hierarchies constructed from similarity orderings in the case where they have been subjected to holdout-based validation and in the case where they have not. These results are given under the heading 'Unoptimized'. For convenience, we copy the results of Table 6 under the heading 'Optimized'.

As in the optimized case, identifying and exploiting variable frontiers in unoptimized clusterings appears to simplify a clustering substantially with no degradation in error rate. Of most interest here, however, is that optimized clusterings are simplified to a substantially greater extent than unoptimized clusterings with no degradation in error rate.

Thus far, we have focused an the criteria of error rate and simplicity, but in many applications, our real interest in simplicity stems from a broader interest in minimizing the expected cost of exploiting a clustering during classification: we expect that simpler clusterings have lower expected classification costs. We can view the various distinctions between unvalidated/validated and unoptimized/optimized clusterings in terms of expected classification cost. Table 8 shows some additional data obtained from our experiments with validation. In particular, the table shows:

**Leaves (L)** The mean number of leaves (over 20 trials) before and after validation (assuming the coarse pruning strategy described in Section 4.2) in both the optimized and unoptimized cases (copied from Table 7).

**EPL** The mean total path length (over 20 trials). The total path length of an unvalidated tree, where each leaf corresponds to a single observation, is the sum of depths of leaves in the tree. In the case of a validated tree, where a leaf may cover multiple observations, the contribution of the leaf to the total path length is the depth of the leaf times the number of observations at that leaf.

**Depth (D)** The average depth of a leaf in the tree, which is computed by $\frac{EPL}{L}$.

**Breadth (B)** The average branching factor of the tree. Given that $B^D = L$, $B = \sqrt[D]{L}$ or $B = m^{\frac{\log_m L}{D}}$ for any $m$.

|  | Unoptimized | | Optimized | |
| --- | --- | --- | --- | --- |
|  | Unvalidated | Validated | Unvalidated | Validated |
| Soybean (small) | | | | |
| Leaves | 18.00 (0.00) | 15.35 (1.81) | 18.00 (0.00) | 13.10 (1.59) |
| EPL | 40.90 (3.64) | 31.90 (6.94) | 54.20 (4.74) | 34.50 (6.49) |
| Depth* | 2.27 | 2.08 | 3.01 | 2.63 |
| Breadth* | 3.57 | 3.72 | 2.61 | 2.66 |
| Cost* | 8.10 | 7.74 | 7.86 | 7.00 |
| Soybean (large) | | | | |
| Leaves | 122.00 (0.00) | 88.55 (4.46) | 122.00 (0.00) | 79.10 (5.80) |
| EPL | 437.20 (34.74) | 280.40 (28.07) | 657.65 (28.38) | 380.65 (43.63) |
| Depth* | 3.58 | 3.17 | 5.39 | 4.81 |
| Breadth* | 3.82 | 4.11 | 2.44 | 2.48 |
| Cost* | 13.68 | 13.03 | 13.15 | 11.93 |
| House | | | | |
| Leaves | 174.00 (0.00) | 68.95 (8.15) | 174.00 (0.00) | 49.10 (7.18) |
| EPL | 664.65 (41.16) | 196.20 (35.32) | 1005.10 (27.42) | 217.25 (39.75) |
| Depth* | 3.82 | 2.85 | 5.78 | 4.42 |
| Breadth* | 3.86 | 4.42 | 2.44 | 2.41 |
| Cost* | 14.75 | 12.60 | 14.10 | 10.65 |
| Mushroom | | | | |
| Leaves | 400.00 (0.00) | 145.50 (20.64) | 400.00 (0.00) | 96.30 (11.79) |
| EPL | 2238.20 (123.63) | 660.90 (117.86) | 2608.85 (56.01) | 503.40 (72.22) |
| Depth* | 5.60 | 4.54 | 6.52 | 5.23 |
| Breadth* | 2.92 | 3.00 | 2.51 | 2.39 |
| Cost* | 16.35 | 13.62 | 16.37 | 12.50 |

Table 8: Cost characteristics of unoptimized and optimized clustering before and after validation. Average and standard deviations over 20 trials. Characteristics that are *ed are computed from the mean values of 'Leaves' and EPL.

**Cost (C)** The expected cost of classifying an observation from the root to a leaf in terms of the number of nodes (clusters) examined during classification. At each level we examine each cluster and select the best. Thus, cost is the product of the number of levels and the number of clusters per level. So $C = B \times D$.

Table 8 illustrates that the expected cost of classification is less for optimized clusterings than for unoptimized clusterings in both the unvalidated and validated cases. However, these results should be taken with a grain of salt, and not simply because they are estimated values. In particular, we have expressed cost in terms of the expected number of nodes that would need to be examined during classification. An implicit assumption is that cost of examination is constant across nodes. In fact, the cost per examination *roughly is constant* (per domain) across nodes in our implementation and many others: at each node, all variables are examined. Consider that by this measure of cost, the least cost (unvalidated) clustering is one that splits the observations in thirds at each node, thus forming a balanced ternary tree, regardless of the form of structure in the data.

Of course, if such a tree does not reasonably capture structure in data, then we might expect this to be reflected in error rate and/or post-validation simplicity. Nonetheless, there are probably better measures of cost available. In particular, Gennari (1989) observed that when classifying an observation, evaluating the objective function over a proper subset of variables is often sufficient to categorize the observation relative to the same cluster that would have been selected if evaluation had occurred over all variables. Under ideal circumstances, when clusters of a partition are well separated (decoupled), testing a very few 'critical' variables may be sufficient to advance classification.

Gennari implemented a *focusing* algorithm that sequentially evaluated the objective function over the variables, one additional variable at a time from most to least 'critical', until a categorization with respect to one of the clusters could be made unambiguously. Using Gennari's procedure, examination cost is not constant across nodes.[12] Carnes and Fisher (Fisher, Xu, Carnes, Reich, Fenves, Chen, Shiavi, Biswas, & Weinberg, 1993) adapted Gennari's procedure to good effect in a diagnosis task, where the intent was to minimize the number of probes necessary to diagnose a fault. While Gennari offers a principled focusing strategy that can be used in conjunction with an objective function, the general idea of focusing on selected features during classification can be traced back to Unimem (Lebowitz, 1982, 1987) and Cyrus (Kolodner, 1983).

The results of Table 8 illustrate the form of an expected classification-cost analysis, but we might have also measured cost as time directly using a test set. In fact, comparisons between the time requirements of sorting in the random and similarity ordering conditions of Tables 4 and 5 suggest cost differences between good and poor clusterings in terms of time as well. Regardless of the form of analysis, however, it seems desirable that one express branching factor and cost in terms of the number of variables that need be tested assuming a focusing strategy such as Gennari's. It is likely that this will tend to make better distinctions between clusterings.

## 5.2 Evaluating Objective Functions: Getting the Most Bang for the Buck

The results of Section 5.1 suggest that the $PU$ function is useful in identifying structure in data: clusterings optimized relative to this function were simpler and as accurate as clusterings that were not optimized relative to the function. Thus, $PU$ leads to something reasonable along the error rate and simplicity dimensions, but can other objective functions do a better job along these dimensions? Based on earlier discussion on the limitations of $PU$, notably that averaging $CU$ over the clusters of a partition introduced 'cliffs' in the space of partitions, it is likely that better objective functions can be found. For example, we might consider Bayesian variants like those found in Autoclass (Cheeseman et al., 1988) and Anderson and Matessa's (1991) system, or the closely related MML approach of Snob (Wallace & Dowe, 1994). We do not evaluate alternative measures such as these here, but do suggest a number of other candidates.

---

12. In fact, cost is not constant across observations, even those that are classified along exactly the same path – the number of variables that one need test depends on the observation's values along previously examined variables.

Section 2.1 noted that the $CU$ function could be viewed as a summation over Gini Indices, which measured the collective impurity of variables conditioned on cluster membership. Intuition may be helped further by an information-theoretic analog to $CU$ (Corter & Gluck, 1992):

$$P(C_k) \sum_i \sum_j [P(A_i = V_{ij}|C_k) \log_2 P(A_i = V_{ij}|C_k) - P(A_i = V_{ij}) \log_2 P(A_i = V_{ij})].$$

The information-theoretic analog can be understood as a summation over *information gain* values, where information gain is an often used selection criterion for decision tree induction (Quinlan, 1986): the clustering analog rewards clusters, $C_k$, that maximize the sum of information gains over the individual variables, $A_i$.

Both the Gini and Information Gain measures are often-used bases for selection measures of decision tree induction. They are used to measure the expected decrease in impurity or uncertainty of a class label, conditioned on knowledge of a given variable's value. In a clustering context, we are interested in the decrease in impurity of *each* variable's value conditioned on knowledge of cluster membership – thus, we use a summation over suitable Gini Indices or alternatively, information gain scores. However, it is well known that in the context of decision tree induction, both measures are biased to select variables with more legal values. Thus, various normalizations of these measures or different measures altogether, have been devised. In the clustering adaptation of these measures normalization is also necessary, since $\sum_{k=1}^{N} CU$ alone or its information-theoretic analog will favor a clustering of greatest cardinality, in which the data are partitioned into singleton clusters, one for each observation. Thus, $PU$ normalizes the sum of Gini indices by averaging.

A general observation is that many selection measures used for decision tree induction can be adapted as objective functions for clustering. There are a number of selection measures that suggest themselves as candidates for clustering, in which normalization is more principled than averaging. Two candidates are Quinlan's (1986) Gain Ratio and Lopez de Mantaras' (1991) normalized information gain.[13]

$$\frac{\sum_j P(A_i=V_{ij}) \sum_k [P(C_k|A_i=V_{ij}) \log_2 P(C_k|A_i=V_{ij}) - P(C_k) \log_2 P(C_k)]}{-\sum_j P(A_i=V_{ij}) \log_2 P(A_i=V_{ij})} \quad \text{(Quinlan, 1986)}$$

$$\frac{\sum_j P(A_i=V_{ij}) \sum_k [P(C_k|A_i=V_{ij}) \log_2 P(C_k|A_i=V_{ij}) - P(C_k) \log_2 P(C_k)]}{-\sum_j \sum_k P(C_k \wedge A_i=V_{ij}) \log_2 P(C_k \wedge A_i=V_{ij})} \quad \text{(Lopez de Mantaras, 1991)}$$

From these we can derive two objective functions for clustering:

$$\sum_i \frac{\sum_k P(C_k) \sum_j [P(A_i=V_{ij}|C_k) \log_2 P(A_i=V_{ij}|C_k) - P(A_i=V_{ij}) \log_2 P(A_i=V_{ij})]}{-\sum_k P(C_k) \log_2 P(C_k)}$$

$$\sum_i \frac{\sum_k P(C_k) \sum_j [P(A_i=V_{ij}|C_k) \log_2 P(A_i=V_{ij}|C_k) - P(A_i=V_{ij}) \log_2 P(A_i=V_{ij})]}{-\sum_k \sum_j P(A_i=V_{ij} \wedge C_k) \log_2 P(A_i=V_{ij} \wedge C_k)}$$

---

13. Jan Hajek independently pointed out the relationship between the $CU$ measure and the Gini Index, and made suggestions on when one might select one or another of the normalizations above.

The latter of these clustering variations was defined in Fisher and Hapanyengwi (1993). Our nonsystematic experimentation with Lopez de Mantaras' normalized information gain variant suggests that it mitigates the problems associated with $PU$, though conclusions about its merits must await further experimentation. In general, there are a wealth of promising objective functions based on decision tree selection measures that we might consider. We have described two, but there are others such as Fayyad's (1991) ORT function.

The relationship between supervised and unsupervised measures also has been pointed out in the context of Bayesian systems (Duda & Hart, 1973). Consider AUTOCLASS (Cheeseman et al., 1988), which searches for the most probable clustering, $H$, given the available data set, $D$ – i.e., the clustering with highest $P(H|D) \propto P(D|H)P(H)$. Under independence assumptions made by AUTOCLASS, the computation of $P(D|H)$ includes the easily seen mechanisms of the simple Bayes classifier used in supervised contexts.

We have not compared the proposed derivations of decision tree selection measures or the Bayesian/MML measures of AUTOCLASS and SNOB as yet, but have proposed pattern-completion error rate, simplicity, and classification cost as external, objective criteria that could be used in such comparisons. An advantage of the Bayesian and MML approaches is that, with proper selection of prior biases, these do not require a separate strategy (e.g., resampling) for pruning, and these strategies can be adapted for variable frontier identification. Rather, the objective function used for cluster formation serves to cease hierarchical decomposition as well. Though we know of no experimental studies with the Bayesian and MML techniques along the accuracy and cost dimensions outlined here, we expect that each would perform quite well.

### 5.3 Final Comments on External Validation Criteria

Our proposal of external validation criteria for clustering such as error rate and classification cost stem from a larger, often implicit, but long-standing bias of some in AI that learning systems should serve the ends of some artificial autonomous agent. Certainly, the COBWEB family of systems trace their ancestry to systems such as UNIMEM (Lebowitz, 1982) and CYRUS (Kolodner, 1983) in which autonomous agency was a primary theme, as it was in Fisher (1987a); Anderson and Matessa's (1991) work expresses similar concerns. In short, the view that clustering is a means of organizing a memory of observations for an autonomous agent begs the question of which of the agent's tasks is memory organization intended to support? Pattern completion error rate and simplicity/cost seem to be obvious candidate criteria.

However, an underlying assumption of this article is that these criteria are also appropriate for externally validating clusterings used in data analysis contexts, where the clustering is external to a human analyst, but is nonetheless exploited by the analyst for purposes such as hypothesis generation. Traditional criteria for cluster evaluation in such contexts include measures of intra-cluster cohesion (i.e., observations within the same clusters should be similar) and inter-cluster coupling (i.e., observations in differing clusters should be dissimilar). The criteria proposed in this article and traditional criteria are certainly related. Consider the following derivation of a portion of the category utility measure, which begins with the *expected number* of variable values that will be correctly predicted given that prediction is guided by a clustering $\{C_1, C_2, ..., C_N\}$:

$E(\# \text{ of correct variable predictions}|\{C_1, C_2, ..., C_N\})$
 $= \sum_k P(C_k) E(\# \text{ of correct variable predictions}|C_k)$
 $= \sum_k P(C_k) \sum_i E(\# \text{ of correct predictions of variable } A_i \ |C_k)$
 $= \sum_k P(C_k) \sum_i \sum_j P(A_i = V_{ij}|C_k) E(\# \text{ of times that } V_{ij} \text{ is correct prediction of } A_i|C_k)$
 $= \sum_k P(C_k) \sum_i \sum_j P(A_i = V_{ij}|C_k) \times P(A_i = V_{ij}|C_k)$
 $= \sum_k P(C_k) \sum_i \sum_j P(A_i = V_{ij}|C_k)^2$

The final steps of this derivation assume that a variable value is predicted with probability $P(A_i = V_{ij}|C_k)$ and that with the same probability this prediction is correct – i.e., the derivation of category utility assumes a *probability matching* prediction strategy (Gluck & Corter, 1985; Corter & Gluck, 1992).[14] By favoring partitions that improve prediction along many variables, hierarchical clustering using category utility tends to result in hierarchies with more variable frontiers, as described in Section 4.1, near the top of the clustering; this tends to reduce post-validation classification cost.

Thus, category utility can be motivated as a measure that rewards cohesion within clusters and decoupling across clusters as noted in Section 2.1, or as a measure motivated by a desire to reduce error rate (and indirectly, classification cost). In general, measures motivated by a desire to reduce error rate will also favor cohesion and decoupling; this stems from two aspects of the pattern-completion task (Lebowitz, 1982; Medin, 1983). First, we assign an observation to a cluster based on the known variable values of the observation, which is best facilitated if variable value predictiveness is high across many variables (i.e., clusters are decoupled).[15] Having assigned an observation to a cluster, we use the cluster's definition to predict the values of variables that are not known from the observation's description. This process is most successful when many variables are predictable at clusters (i.e., clusters are cohesive). In fact, designing measures with cohesion and decoupling in mind undoubtedly results in useful clusterings for purposes of pattern completion, whether or not this was the explicit goal of the designer.

If external validation criteria of error rate and cost are well correlated with traditional criteria of cohesion and coupling, then why use the former criteria at all? In part, this stems from an AI and machine learning bias that systems should be designed and evaluated with a specific performance task in mind. In addition, however, a plethora of measures for assessing cohesion and coupling can be found, with each system assessed relative to some variant. This variation can make it more difficult to assess similarities and differences across systems. This article suggests pattern-completion error rate and cost as relatively unbiased alternatives for comparative studies. Inversely, why not use some direct measures of error rate and classification cost (e.g., using holdout) as an 'objective function' to guide search through the space of clusterings? This can be expensive. Thus, we use a cheaply computed objective function that is designed with external error rate and cost evaluation in mind; undoubtedly, such an objective function reflects cohesion and coupling.

---

14. Importantly, prediction with COBWEB is actually performed using a *probability maximizing* strategy – the most frequent value of a variable at a cluster is always predicted. Fisher (1987b) discusses the advantage of constructing clusters with an implicit probability matching strategy, even in cases where these clusters will be exploited with a probability maximizing strategy.
15. The MML and Bayesian approaches of SNOB and AUTOCLASS support probabilistic assignment of observations to clusters, but the importance of decoupling and cohesion remain.

Of course, we have computed error rate and identified variable frontiers given a simplified performance task: each variable was independently masked and predicted over test observations. This is not an unreasonable generic method for computing error rate, but different domains may suggest different computations, since often many variables are simultaneously unknown and/or an analyst may be interested in a subset of the variables. In addition, we have proposed simplicity (i.e., the number of leaves) and expected classification cost as external validation criteria. Section 5.1 suggests that one of the latter criteria is probably necessary, in addition to error rate, to discriminate 'good' and 'poor' clusterings as judged by the objective function. In general, desirable realizations of error rate, simplicity, and cost will likely vary with domain and the interpretation tasks of an analyst.

In short, an analyst's task is largely one of making inferences from a clustering, for which there are error-rate and cost components (i.e., what information can an analyst glean from a clustering and how much work is required on the part of the analyst to extract this information). It is probably not the case that we have expressed these components in precisely the way that they are cognitively-implemented in an analyst. Nonetheless, this article and others (Fisher, 1987a; Cheeseman et al., 1988; Anderson & Matessa, 1991) can be viewed as attempts to formally, but tentatively *describe* an analyst's criteria for cluster evaluation, based on criteria that we might *prescribe* for an autonomous, artificial agent confronted with much the same task.

### 5.4 Other Issues

There are many important issues in clustering that we will not address in depth. One of these is the possible advantage of overlapping clusters (Lebowitz, 1987; Martin & Billman, 1994). We have assumed tree-structured clusterings, which store each observation in more than one cluster, but these clusters are related by a proper subset-of relation as one descends a path in the tree. In many cases, *lattices* (Levinson, 1984; Wilcox & Levinson, 1986; Carpineto & Romano, 1993), or more generally, *directed acyclic graphs* (DAG) may be a better representation scheme. These structures allow an observation to be included in multiple clusters, where one such cluster need not be a subset of another. As such, they may better provide an analyst with multiple perspectives of the data. For example, animals can be partitioned into clusters corresponding to mammals, birds, reptiles, etc., or they may be partitioned into clusters corresponding to carnivores, herbivores, or omnivores. A tree would require that one of these partitions (e.g., carnivore, etc.) be 'subordinate' to the other (e.g., mammals, birds, etc.); Classes of the subordinate partition would necessarily be 'distributed' across descendents (e.g., carnivorous-mammal, omnivorous-mammal, carnivorous-reptile, etc.) of top level clusters, which ideally would represent clusters of the other partition. A DAG allows both perspectives to coexist in relative equality, thus making both perspectives more explicit to an analyst.

We have also assumed that variables are nominally valued. There have been numerous adaptations of the basic *PU* function, other functions, and discretization strategies to accommodate numeric variables (Michalski & Stepp, 1983a, 1983b; Gennari et al., 1989; Reich & Fenves, 1991; Cheeseman et al., 1988; Biswas et al., 1994). The basic sorting procedure and the iterative optimization techniques can be used with data described in whole or part by numerically-valued variables regardless of which approach one takes. The identification of numeric variable frontiers using holdout can be done by using the mean value for a variable at a node for generating predictions, and identifying a variable's frontier as the set of clusters that collectively minimize a measure of error such as mean-squared error.

## 6. Concluding Remarks

We have partitioned the search through the space of hierarchical clusterings into three phases. These phases, together with an opinion of their desirable characteristics from a data analysis standpoint, are (1) inexpensive generation of an initial clustering that suggests the form of structure in data (or its absence), (2) iterative optimization (perhaps in background) for clusterings of better quality, and (3) retrospective simplification of generated clusterings. We have evaluated three iterative optimization strategies that operate independent of objective function. All of these, to varying degrees, are inspired by previous research, but hierarchical redistribution appears novel as an iterative optimization technique for clustering; it also appears to do quite well.

Another novel aspect of this work is the use of resampling as a means of validating clusters and of simplifying hierarchical clusterings. The experiments of Section 5 indicate that optimized clusterings provide greater data compression than do unoptimized clusterings. This is not surprising, given that $PU$ compresses data in some reasonable manner; whether it does so 'optimally' though is another issue.

We have made several recommendations for further research.

1. We have suggested experiments with alternative objective functions, including Bayesian and MML measures, and some that are inspired by variable-selection measures of decision tree induction.

2. There may be cost and quality benefits to applying optimization strategies at intermittent points during hierarchical sorting.

3. The holdout method of identifying variable frontiers and pruning suggests a strategy akin to n-fold-cross validation that clusters over all the data, while still identifying variable frontiers and facilitating pruning.

4. Analyses of classification cost for purposes of external validation are probably best expressed in terms of the expected number of variables using a focusing method such as Gennari's.

In sum, this paper has proposed criteria for internal and external validation, and has made experimental comparisons between various approaches along these dimensions. Ideally, as researchers explore other objective functions, search control strategies, and pruning techniques, the same kind of experimental comparisons (particularly along external criteria such as error rate, simplicity, and classification cost) that are *de rigueur* in comparisons of supervised systems, will become more prominent in unsupervised contexts.

## Acknowledgements

I thank Sashank Varma, Arthur Nevins, and Diana Gordon for comments on the paper. The reviewers and editor supplied extensive and helpful comments. This work was supported by grant NAG 2-834 from NASA Ames Research Center. A very abbreviated discussion of some of this article's results appear in Fisher (1995), published by AAAI Press.